\documentclass{article}
\usepackage{amsmath,amsthm, amssymb,amsfonts}
\usepackage{spconf,graphicx}
\usepackage{enumitem}
\usepackage{diagbox}
\usepackage{caption}

\title{Multiple-hypothesis CTC-based semi-supervised adaptation of end-to-end speech recognition}
\name{Cong-Thanh Do$^{1}$, Rama Doddipatla$^{1}$, and Thomas Hain$^{2}$}
\address{$^{1}$Toshiba Cambridge Research Laboratory, Cambridge, UK \\
$^{2}$The University of Sheffield, Sheffield, UK}

\begin{document}
\setlength{\abovedisplayskip}{1pt}
\ninept
\maketitle
\begin{abstract}
This paper proposes an adaptation method for end-to-end speech recognition. In this method, multiple automatic speech recognition (ASR) 1-best hypotheses are integrated in the computation of the connectionist temporal classification (CTC) loss function. The integration of multiple ASR hypotheses helps alleviating the impact of errors in the ASR hypotheses to the computation of the CTC loss when ASR hypotheses are used. When being applied in semi-supervised adaptation scenarios where part of the adaptation data do not have labels, the CTC loss of the proposed method is computed from different ASR 1-best hypotheses obtained by decoding the unlabeled adaptation data. Experiments are performed in clean and multi-condition training scenarios where the CTC-based end-to-end ASR systems are trained on Wall Street Journal (WSJ) clean training data and CHiME-4 multi-condition training data, respectively, and tested on Aurora-4 test data. The proposed adaptation method yields 6.6\% and 5.8\% relative word error rate (WER) reductions in clean and multi-condition training scenarios, respectively, compared to a baseline system which is adapted with part of the adaptation data having manual transcriptions using back-propagation fine-tuning.  
\end{abstract}
\section{Introduction}
\label{sec:intro}

Mismatch between training and test data is common when using automatic speech recognition (ASR) systems in realistic conditions. Among other robustness methods, adaptation algorithms developed for ASR aim at alleviating this mismatch. Adapting large and complex models, especially deep neural network (DNN)-based models, is challenging with typically a small amount of adaptation (target) data and without explicit supervision \cite{bell2020}.

Adaptation algorithms use adaptation data, which should be matched to the target test data, to adapt the trained ASR system and close the gap between training and test. The transcriptions, or labels, of the adaptation data are required in supervised adaptation. However, manual transcriptions are not always available because obtaining these transcriptions for a large amount of data is costly. When manual transcriptions are not available, ASR hypotheses, or ``pseudo-labels", can be used in the place of manual transcriptions. The ASR hypotheses are obtained by decoding the adaptation data using the trained (non-adapted) system. When ASR hypotheses are used, inaccurate information is present because the automatic transcriptions are typically not free of errors.

End-to-end speech recognition uses a single neural network architecture within the deep learning framework to perform speech-to-text task. In the training of end-to-end speech recognition systems, the need for having prior alignments between acoustic frames and output symbols is eliminated thanks to the use of training criteria such as the attention mechanism \cite{chorowski2015} or the connectionist temporal classification (CTC) loss function \cite{graves2014}.

Connectionist temporal classification (CTC) is the process of automatically labeling unsegmented data sequences using a neural network \cite{graves2006}. The training of a neural network using the CTC loss function thus does not require prior alignments between the input and target sequences. In the training of neural network using CTC loss and characters as output symbols, for a given transcription of the input sequence, there are as many possible alignments as there are different ways of separating the characters with blanks. As the exact character sequence, derived from the transcription, corresponding to the input sequence is not known, the sum over all possible character sequences is performed \cite{graves2014}. In semi-supervised or unsupervised adaptation where ASR hypotheses are used, the computation of the CTC loss could be unfavorably affected because there are errors in the transcriptions which are in essence the ASR hypotheses.

In this paper, we propose an adaptation method for CTC-based end-to-end speech recognition in which the impact of errors in the transcriptions to the CTC loss computation is alleviated by combining CTC losses computed from different ASR 1-best hypotheses. In the present paper, the ASR 1-best hypotheses are obtained by using ASR systems with different acoustic features to decode the unlabeled adaptation data. We show the effectiveness of the proposed adaptation method in semi-supervised adaptation scenarios where the CTC-based end-to-end speech recognition systems are trained either on clean training data from the Wall Street Journal (WSJ) corpus \cite{paul1992} or on multi-condition training data of the CHiME-4 corpus \cite{vincent2017}, while evaluating on the test data of Aurora-4 corpus \cite{parihar2002}.

The paper is organized as follows. Section \ref{sec:related_works} presents related works. The proposed adaptation method using multiple ASR hypotheses and CTC losses combination is introduced in section \ref{sec:adaptation_method}. Sections \ref{sec:asr_system_data} and \ref{sec:adaptation_experiments} present about ASR systems training and adaptation experiments, respectively. Results are presented in section \ref{sec:results}. Finally, section \ref{sec:conclusion} concludes the paper.

\vspace{-5pt}
\section{Related works}
\label{sec:related_works}

Adaptation of end-to-end speech recognition has been investigated in a number of studies \cite{samarakoon2018, li_slt2018, ochiai2018, delcroix2018, meng2019, tsunoo2019, sari_icassp2020, do_eusipco2020, ding2020}. In \cite{li_slt2018}, adaptation of the end-to-end model was achieved by introducing Kullback-Leibler divergence (KLD) regularization and multi-task learning (MTL) criterion into the CTC loss function. The training criteria are the linear combination of the standard CTC loss and the KLD or the MTL criterion. Multiple hypotheses were previously used in cross-system acoustic model adaptation where the transcriptions for adaptation were generated by several systems, which were built with various phoneme sets or acoustic front-ends \cite{giuliani2007, stueker2006}.

In the present work, a new loss function created by combining the CTC losses computed from different ASR 1-best hypotheses is used during adaptation. The ASR 1-best hypotheses are obtained by decoding the unlabeled adaptation data with ASR systems using different acoustic features.

\section{Proposed adaptation method}
\label{sec:adaptation_method}

\subsection{Training of CTC-based end-to-end speech recognition}
\label{sec:ctc_training}

Given a $T$-length acoustic feature vector sequence $X = \{\mathbf{x}_t \in \mathbb{R}^d | t = 1,...,T\}$, where $\mathbf{x}_t$ is a $d$-dimensional feature vector at frame $t$, and a transcription $C = \{c_l \in \mathcal{U}| l = 1,...,L\}$ which consists of $L$ characters, where $\mathcal{U}$ is a set of distinct characters, during the training of the neural network the standard CTC loss function $L_{CTC}$ is defined as follows:

\begin{equation}
\label{equ:ctc_loss}
L_{CTC} = -\log P_{\theta} (C|X),
\end{equation}

\noindent where $\theta$ are the network parameters. The network is trained to minimize $L_{CTC}$. In equation (\ref{equ:ctc_loss}), $C$ is the transcription of $X$ which can be either a manual transcription or an ASR hypothesis. In the present work, the ASR systems are trained using manual transcriptions in supervised training mode. The convolutional neural network (CNN) \cite{lecun1995} - bidirectional long short-term memory (BLSTM) \cite{hochreiter1997} architecture is used.

The CTC loss function in equation (\ref{equ:ctc_loss}) can be computed thanks to the introduction of the CTC path $a$ which forces the output character sequence to have the same length as the input feature sequence by adding blank as an additional label and allowing repetition of labels \cite{graves2014}. The CTC loss $L_{CTC}$ is thus computed by integrating over all possible CTC paths $\mathcal{B}^{-1}(C)$ expanded from $C$:

\begin{equation}
\label{equa:ctc_paths}
L_{CTC} = -\log P_{\theta} (C|X) = -\log\sum_{a \in \mathcal{B}^{-1}(C)}P_{\theta} (a|X).
\end{equation}

\subsection{Multiple-hypothesis CTC-based adaptation}
\label{sec:ctc_adaptation}

Given adaptation data, among other adaptation methods mentioned in section \ref{sec:related_works}, the CTC-based end-to-end speech recognition system can be adapted by using back-propagation algorithm \cite{rumelhart1986} to fine-tune the trained neural network \cite{do_eusipco2020}. During the minimization of the CTC loss function using stochastic gradient descent \cite{ruder2016}, the parameters of the neural network are updated. When the manual transcriptions of the adaptation data are not available, ASR 1-best hypotheses obtained by using the trained neural network to decode the adaptation data can be used in the adaptation process. 

In this paper, we propose to integrate multiple ASR 1-best hypotheses in the computation of the CTC loss function during adaptation, when the manual transcriptions are not available, as follows:

\begin{equation}
\label{equa:multiple_hypothesis_ctc}
L^{*}_{CTC} = -\left( \sum^{N}_{i=1} \log P_{\theta} (\widehat{C}_i|X) \right),
\end{equation}

\noindent where $\widehat{C}_i, i = 1, ..., N$ are the 1-best hypotheses obtained by decoding the unlabeled adaptation data using $N$ different trained neural networks. By combining multiple 1-best hypotheses in the computation of the CTC loss, the impact of the errors in the ASR hypotheses to the computation of the CTC loss function could be alleviated. Using property of the logarithm, the equation (\ref{equa:multiple_hypothesis_ctc}) can be rewritten as follows:

\begingroup
\scriptsize
\begin{equation}
\label{equa:ctc_product}
L^{*}_{CTC} = - \log \prod^{N}_{i=1} P_{\theta} (\widehat{C}_i|X) = - \log\prod^{N}_{i=1}\left(\sum_{a_i \in \mathcal{B}^{-1}(\widehat{C}_i)}P_{\theta} (a_i|X)\right),
\end{equation}
\endgroup


\noindent where $a_i$ is a CTC path linking the 1-best hypothesis $\widehat{C}_i$ and the acoustic feature sequence $X$.


In the computation of the new CTC loss $L^{*}_{CTC}$ in the present paper, different ASR 1-best hypotheses are obtained by decoding the adaptation data with different ASR systems. Different ASR hypotheses could be obtained by other means, for instance by using N-best hypotheses from one decoding. This possibility is not explored in the present paper. Also, no confidence-based filtering \cite{bell2020} is applied on the ASR hypotheses. In the experiments of the present paper, the use of two systems ($N=2$) is explored.


\subsection{Analysis}
\label{sec:anslysis}

We analyze the new loss function $L^{*}_{CTC}$ for the simplified case where two 1-best hypotheses are used. The equation (\ref{equa:ctc_product}) becomes: 

\begingroup
\scriptsize
\begin{equation}
\label{equa:ctc_product_2}
L^{*}_{CTC} = - \log \left[ \left(\sum_{a_i \in \mathcal{B}^{-1}(\widehat{C}_1)}P_{\theta} (a_i|X)\right)\left(\sum_{b_j \in \mathcal{B}^{-1}(\widehat{C}_2)}P_{\theta} (b_j|X)\right)\right],
\end{equation}
\endgroup

\noindent where $a_i$ and $b_j$ are ones of the CTC paths linking the 1-best hypotheses $\widehat{C}_1$ and $\widehat{C}_2$, respectively, with the acoustic feature sequence $X$. From equation (\ref{equa:ctc_product_2}), it can be seen that a probability $P_{\theta} (a_i|X)$, computed by using the CTC path $a_i$, would be multiplied with all the probabilities $P_{\theta} (b_j|X), b_j \in \mathcal{B}^{-1}(\widehat{C}_2)$. This weighting, based on the probabilities computed from different CTC paths in $\mathcal{B}^{-1}(\widehat{C}_2)$, could possibly alleviate the impact of uncertainty in the CTC paths $a_i \in \mathcal{B}^{-1}(\widehat{C}_1)$, caused by transcription errors in $\widehat{C}_1$, to the computation of the CTC loss $L^{*}_{CTC}$.

\vspace{-5pt}
\section{Speech recognition system and data}
\label{sec:asr_system_data}

The effectiveness of the proposed adaptation method is evaluated in semi-supervised adaptation scenarios where only part of the adaptation data have manual transcriptions. This scenario is popular when manual transcriptions can be obtained only for a small amount of adaptation data instead of total amount of adaptation data, to reduce the cost. The end-to-end ASR systems are trained using the standard CTC loss function (see equation (\ref{equ:ctc_loss})). The proposed CTC loss function $L^{*}_{CTC}$ is used only in the adaptation using the proposed multiple-hypothesis CTC-based adaptation method. Other adaptations use the standard CTC loss function as in equation (\ref{equ:ctc_loss}).

\subsection{CTC-based end-to-end speech recognition systems}
\label{sec:e2e_systems}

\subsubsection{Acoustic features}
\label{sec:acoustic_features}

CNN-BLSTM neural network architecture is trained with CTC loss to map acoustic feature sequences to character sequences. A baseline system is trained by using 40-dimensional log-Mel filter-bank (FBANK) features \cite{mohamed2012} as acoustic features. The FBANK features are augmented with 3 dimensional pitch features \cite{watanabe2018, povey2011}. Delta and acceleration features are appended to the static features. The feature extraction of the baseline system was performed by using the standard feature extraction recipe of Kaldi toolkit \cite{povey2011}.

To have additional ASR hypotheses, another system is trained to decode the unlabeled adaptation data. The system is trained by using 40-dimensional subband temporal envelope (STE) features \cite{do2017_interspeech} together with 3-dimensional pitch features. Similar to the system trained with FBANK features, the delta and acceleration features are included. STE features track energy peaks in perceptual frequency bands which reflect the resonant properties of the vocal tract. These features have been shown to be on par with the standard FBANK features in various speech recognition scenarios \cite{doddipatla2018, do2019}. FBANK and STE features are also complementary to each other and combining the systems using these features yielded significant WER reductions compared to single system \cite{do2017_interspeech, doddipatla2018, do2019}.

\vspace{-5pt}
\subsubsection{Neural network architecture}
\label{sec:architecture}

The neural network architecture for end-to-end ASR systems is made up of initial layers of the VGG net architecture (deep CNN) \cite{simonyan2015} followed by a 6-layer pyramid BLSTM (BLSTM with subsampling \cite{watanabe2018}). We use a 6-layer CNN architecture which consists of two consecutive 2D convolutional layers followed by one 2D Max-pooling layer, then another two 2D convolutional layers followed by one 2D max-pooling layer. The 2D filters used in the convolutional layers have the same size of 3$\times$3. The max-pooling layers have patch of 3$\times$3 and stride of 2$\times$2. The 6-layer BLSTM has 1024 memory blocks in each layer and direction, and linear projection is followed by each BLSTM layer. The subsampling factor performed by the BLSTM is 4 \cite{watanabe2018}. During decoding, CTC score is used in a one-pass beam search algorithm \cite{watanabe2018}. The beam width is set to 20. Training and decoding are performed using the ESPnet toolkit \cite{watanabe2018}.

\vspace{-5pt}
\subsection{Data}

\subsubsection{Clean training data}
\label{sec:clean_training}

WSJ is a corpus of read speech \cite{paul1992}. All the speech utterances are sampled at 16 kHz and are fairly clean. The WSJ's standard training set \footnotesize \texttt{train\_si284} \normalsize consists of around 81 hours of speech. During training, the standard development set \footnotesize \texttt{test\_dev93}, \normalsize which consists of around 1 hour of speech, is used for cross-validation.

\vspace{-5pt}
\subsubsection{Multi-condition training data}
\label{sec:multi_condition_training}

The multi-condition training data of CHiME-4 corpus \cite{vincent2017} consists of around 189 hours of speech, in total. The CHiME-4 multi-condition training data consists of the clean speech utterances from WSJ training corpus and simulated and real noisy data. The real data consists of 6-channel recordings of utterances from WSJ corpus spoken in four environments: caf\'e, street junction, public transport (bus), and pedestrian area. The simulated data was constructed by mixing WSJ clean utterances with the environment background recordings from the four mentioned environments. All the data were sampled at 16 kHz. Audio recorded from all the microphone channels are included in the CHiME-4 multi-condition training data, named \footnotesize \texttt{tr05\_multi\_noisy\_si284} \normalsize in the ESPnet CHiME-4 recipe. The \footnotesize \texttt{dt05\_multi\_isolated\_1ch\_track} \normalsize set was used for cross-validation during training.

\vspace{-10pt}
\subsubsection{Test and adaptation data}
\label{sec:test}

Test and adaptation sets are created from the test sets of the Aurora-4 corpus \cite{parihar2002}. The Aurora-4 corpus has 14 test sets which were created by corrupting two clean test sets, recorded by a primary Sennheiser microphone and a secondary microphone, with six types of noises: airport, babble, car, restaurant, street, and train, at 5-15 dB SNRs. The two clean test sets were also included in the 14 test sets. There are 330 utterances in each test set. The noises in Aurora-4 are different from those in the CHiME-4 multi-condition training data. In this work, the .wv1 data \cite{parihar2002} from 7 test sets created from the clean test set recorded by the primary Sennheiser microphone are used to create test and adaptation sets.

From 2310 utterances taken from the 7 test sets of .wv1 data, a test set of 1400 utterances (approx. 2.8 hours of speech), a labeled adaptation set of 300 utterances (approx. 36 minutes), and an unlabeled adaptation set of 610 utterances (approx. 1.2 hours) are separated. The selection of the utterances in the three sets are random. The utterances in the three sets are not overlapped. These sets are used for testing and adaptation in both clean training and multi-condition training scenarios.

\vspace{-5pt}
\section{Adaptation experiments}
\label{sec:adaptation_experiments}

\begin{figure}[ht]
	\centering
		\includegraphics[width=1.0\columnwidth]{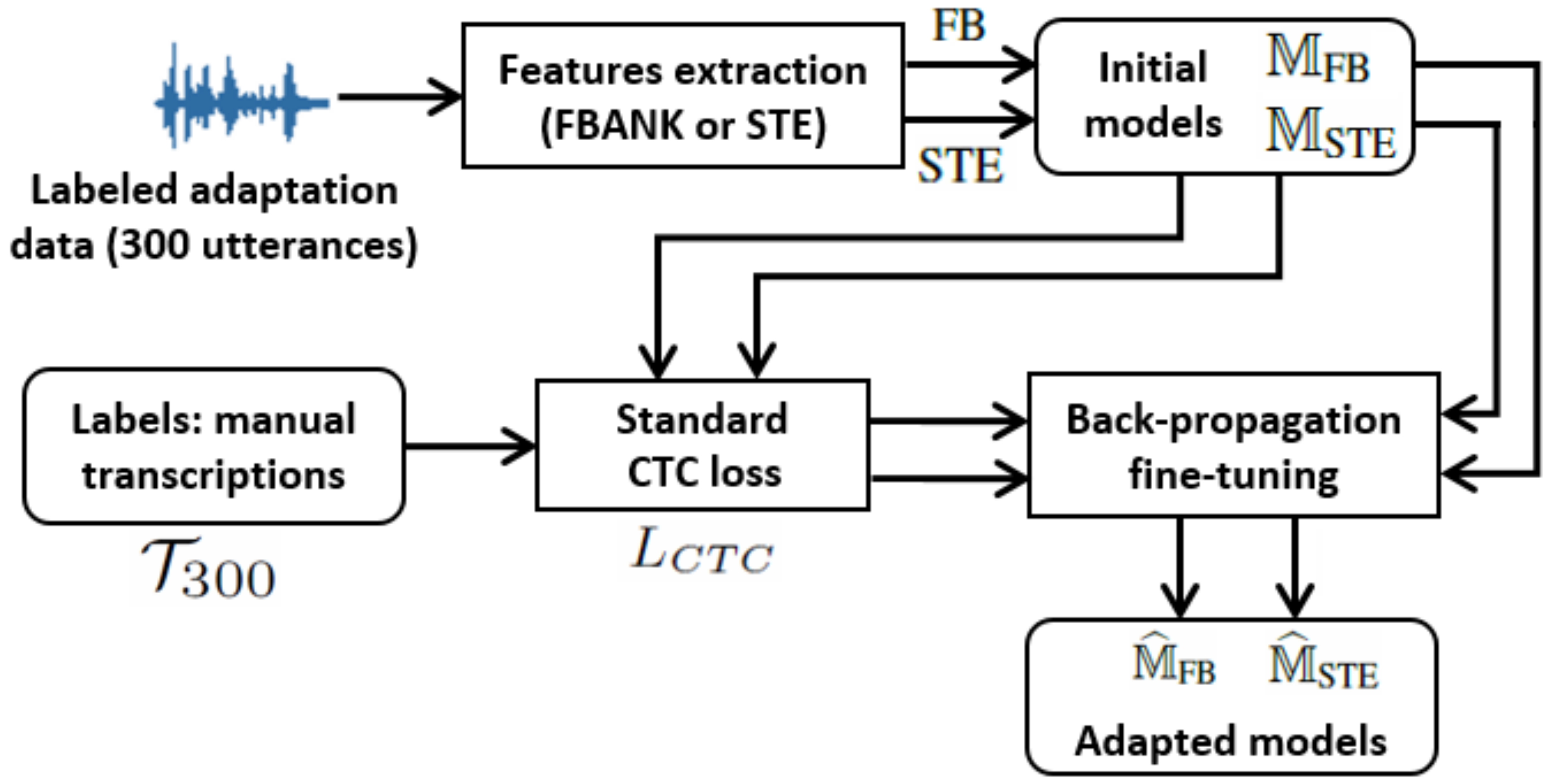}
	\caption{\label{fig:supervised_adapt} Supervised adaptation of initial models $\mathbb{M}_{\text{FB}}$ and $\mathbb{M}_{\text{STE}}$ using the 300-utterance set with manual transcriptions $\mathcal{T}_{300}$. The models can be trained either on clean or multi-condition training data.}
\end{figure}

Let $\mathbb{M}_{\text{FB}}$ and $\mathbb{M}_{\text{STE}}$ be the end-to-end models trained with FBANK and STE features, respectively, on the clean or multi-condition training data, the semi-supervised adaptation experiment is performed as follows (in this section, for the sake of clarity, notations for clean and multi-condition training data are not included):
\begin{itemize}[leftmargin=*]
\item First the back-propagation algorithm is used to fine-tune the models $\mathbb{M}_{\text{FB}}$ and $\mathbb{M}_{\text{STE}}$ in supervised mode using the labeled adaptation set of 300 utterances to obtain the adapted model $\widehat{\mathbb{M}}_{\text{FB}}$ and $\widehat{\mathbb{M}}_{\text{STE}}$, respectively (see Figure \ref{fig:supervised_adapt}). This step is done to make use of the available labeled adaptation data and to reduce further the WERs of the ASR systems.
\item The models $\widehat{\mathbb{M}}_{\text{FB}}$ and $\widehat{\mathbb{M}}_{\text{STE}}$ are subsequently used to decode the unlabeled adaptation set of 610 utterances. Assume that $\mathcal{H}^{\textup{FB}}_{610}$ and $\mathcal{H}^{\textup{STE}}_{610}$ are the sets of 1-best hypotheses obtained from these decoding and $\mathcal{T}_{300}$ is the set of manual transcriptions available for the 300 utterances set, we group the 300-utterance and 610-utterance sets to create an adaptation set of 910 utterances whose labels could be either $\mathcal{T}_{300} \cup \mathcal{H}^{\textup{FB}}_{610}$ or $\mathcal{T}_{300} \cup \mathcal{H}^{\textup{STE}}_{610}$.
\item Finally, the 910-utterance set is used to adapt the model $\mathbb{M}_{\text{FB}}$, which is the initial model, using back-propagation algorithm to obtain the semi-supervised adapted model $\widetilde{\mathbb{M}}_{\text{FB}}$.
\end{itemize}


\begin{figure}[ht]
	\centering
		\includegraphics[width=1.0\columnwidth]{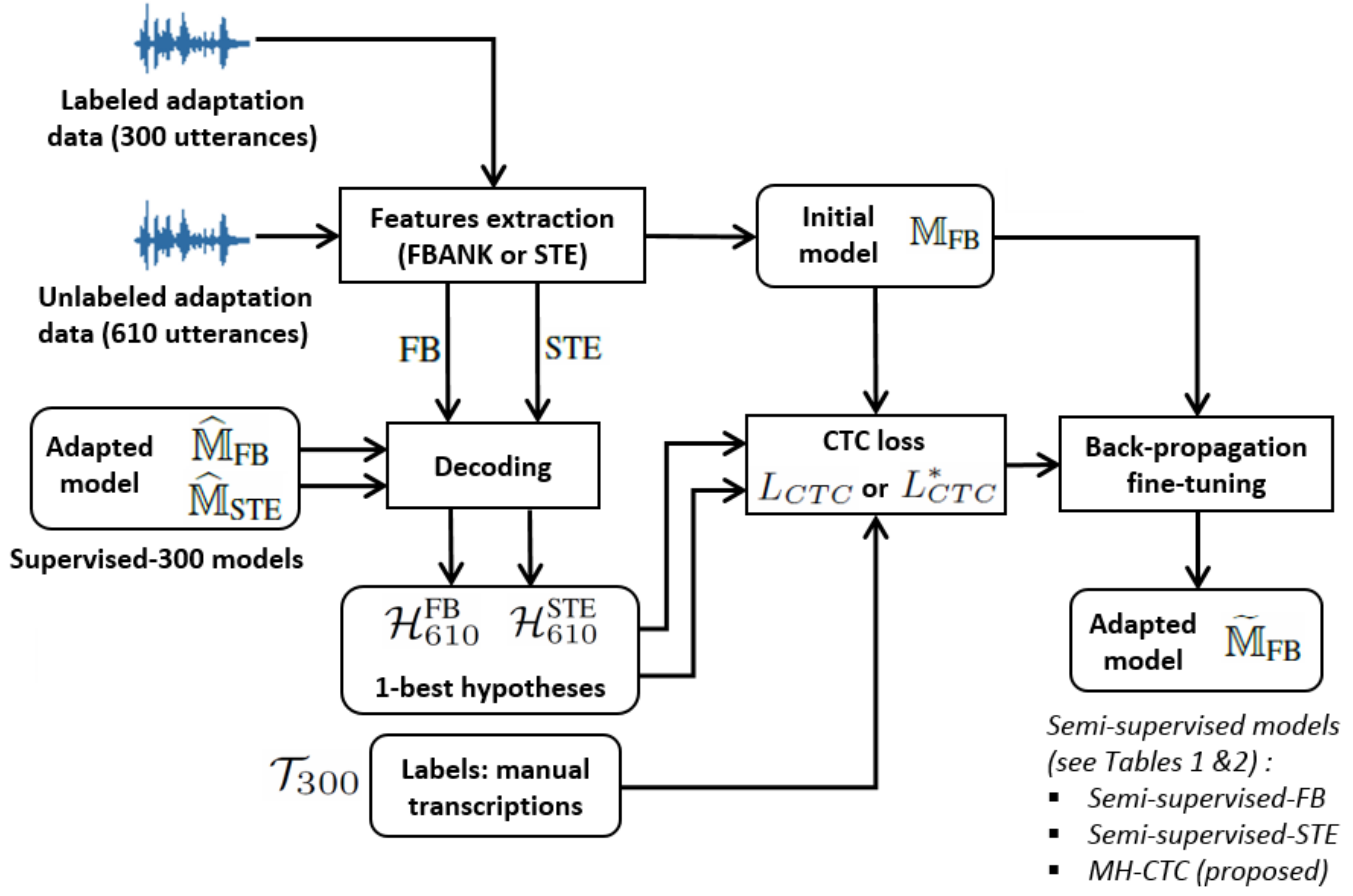}
	\caption{\label{fig:semi_supervised_adapt} Semi-supervised adaptations using the 910-utterance adaptation set, of which the labels include the manual transcriptions $\mathcal{T}_{300}$ and one of the sets of 1-best hypotheses, $\mathcal{H}^{\textup{FB}}_{610}$ and $\mathcal{H}^{\textup{STE}}_{610}$, or both.\vspace{-15pt}}
\end{figure}

The 910-utterance adaptation set in which 610 utterances do not have manual transcriptions is used to adapt the initial FBANK-based system in semi-supervised mode since only 300 utterances have manual transcriptions. The conventional semi-supervised adaptation using the 910-utterance adaptation set can be done with the labels from $\mathcal{T}_{300}$ and, either $\mathcal{H}^{\textup{FB}}_{610}$ or $\mathcal{H}^{\textup{STE}}_{610}$. This adaptation uses the standard CTC loss $L_{CTC}$. The proposed multiple-hypothesis CTC-based adaptation method, denoted as MH-CTC, uses the $\mathcal{T}_{300}$ manual transcriptions and both sets of 1-best hypotheses, $\mathcal{H}^{\textup{FB}}_{610}$ and $\mathcal{H}^{\textup{STE}}_{610}$. This adaptation used the $L^{*}_{CTC}$ loss. These semi-supervised adaptation experiments are depicted in Figure \ref{fig:semi_supervised_adapt}.

The referenced performance which can be considered as an upper bound performance for all the mentioned adaptation methods is that obtained with the supervised adaptation where all 910 utterances have manual transcriptions $\mathcal{T}_{910}$. During adaptation, the learning rate is kept unchanged compared to that used during training because this configuration yields better performance than using different learning rates during training and adaptation. On the other hand, the 1-best hypotheses are obtained after one pass of decoding.

\vspace{-5pt}
\section{Results}
\label{sec:results}

\vspace{-5pt}
\subsection{Clean training}
\label{sec:results_clean_training}

In the scenario where the systems are trained on the WSJ clean training data and tested on the test set consisting of 1400 Aurora-4 utterances, the initial systems which use the models $\mathbb{M}_{\text{FB}}$ and $\mathbb{M}_{\text{STE}}$, respectively, have WERs of 55.2\% and 60.3\%, respectively. The results of applying different adaptation methods to the FBANK-based system are shown in Table \ref{tab:clean_training}. Adapting the initial FBANK-based and STE-based systems with the labeled adaptation set of 300 utterances reduces the WERs of these systems measured on the 1400-utterance test set to 27.2\% and 24.5\%, respectively. The corresponding WERs measured on the 610-utterance unlabeled adaptation set are 29.1\% and 25.6\%, respectively. 

Supervised adaptation using the 300-utterance adaptation set with manual transcriptions $\mathcal{T}_{300}$ is used as the baseline. It can be observed from Table \ref{tab:clean_training}, that, the proposed multiple-hypothesis CTC-based adaptation method yields 6.6\% relative WER reduction compared to the baseline. In contrast, the two conventional semi-supervised adaptations which use both manual transcriptions and one of the sets of 1-best hypotheses, $\mathcal{H}^{\textup{FB-C}}_{610}$ or $\mathcal{H}^{\textup{STE-C}}_{610}$, do not yield WER reduction compared to the FBANK-based baseline system.

\subsection{Multi-condition training}
\label{sec:results_multi_condition_training}

The experiments in the clean training scenario are repeated for the multi-condition training scenario. When being trained on multi-condition training data of CHiME-4 and tested on the 1400-utterance test set from Aurora-4, the initial CTC-based end-to-end ASR systems using FBANK and STE features have WERs of 31.0\% and 33.8\%, respectively. Adapting the initial FBANK-based and STE-based systems with the labeled adaptation set of 300 utterances reduces the WERs of these systems measured on the 1400-utterance test set to 17.2\% and 17.3\%, respectively. The corresponding WERs which are measured on the 610-utterance unlabeled adaptation set are 18.3\% and 18.9\%, respectively. Results of the adaptation experiments in this scenario are shown in Table \ref{tab:multi_condition_training}. Similar to in the clean training scenario, the proposed adaptation method (MH-CTC) yields 5.8\% relative WER reduction compared to the baseline. The semi-supervised adaptations using single 1-best hypotheses $\mathcal{H}^{\textup{FB-M}}_{610}$ or $\mathcal{H}^{\textup{STE-M}}_{610}$ together with the manual transcriptions $\mathcal{T}_{300}$ do not yield WER reduction compared to the baseline.

\begin{table}[t]
\begin{center}
\captionof{table}{Adaptation of the FBANK-based ASR system trained on WSJ clean training set with different adaptation methods. $\mathcal{H}^{\textup{FB-C}}_{610}$ and $\mathcal{H}^{\textup{STE-C}}_{610}$ are obtained in the decoding using clean training models.}
\label{tab:clean_training}
\begin{tabular}{|l|c|l|c|}
 \hline
\footnotesize Adaptation method & \footnotesize \# Utts. & \footnotesize Adapt. data's labels & \footnotesize WER 
 \\\hline\hline
 \footnotesize No adapt. (initial model) & \footnotesize N/A & \footnotesize N/A & \footnotesize 55.2\\
\hline
 \footnotesize Supervised-300 \scriptsize (baseline) & \footnotesize 300 & \scriptsize $\mathcal{T}_{300}$ & \footnotesize 27.2\\
\hline
 \footnotesize Semi-supervised-FB & \footnotesize 910 & \scriptsize $\mathcal{T}_{300} \cup \mathcal{H}^{\textup{FB-C}}_{610}$ & \footnotesize 28.4\\
\hline
 \footnotesize Semi-supervised-STE & \footnotesize 910 & \scriptsize $\mathcal{T}_{300} \cup \mathcal{H}^{\textup{STE-C}}_{610}$ & \footnotesize 27.4\\
\hline
 \footnotesize MH-CTC (proposed) & \footnotesize 910 & \scriptsize $\mathcal{T}_{300} \cup \mathcal{H}^{\textup{FB-C}}_{610} \cup \mathcal{H}^{\textup{STE-C}}_{610}$ & \footnotesize \textbf{25.4}\\
\hline
 \footnotesize Supervised-910 & \footnotesize 910 & \footnotesize $\mathcal{T}_{910}$ & \footnotesize 13.2\\
\hline
\end{tabular}
\end{center}
\normalsize

\vspace{-10pt}
\begin{center}
\captionof{table}{Adaptation of the FBANK-based ASR system trained on CHiME-4 multi-condition training set with different adaptation methods. $\mathcal{H}^{\textup{FB-M}}_{610}$ and $\mathcal{H}^{\textup{STE-M}}_{610}$ are obtained in the decoding using multi-condition training models.}
\label{tab:multi_condition_training}
\begin{tabular}{|l|c|l|c|}
 \hline
\footnotesize Adaptation method & \footnotesize \# Utts. & \footnotesize Adapt. data's labels & \footnotesize WER 
 \\\hline\hline
 \footnotesize No adapt. (initial model) & \footnotesize N/A & \footnotesize N/A & \footnotesize 31.0\\
\hline
 \footnotesize Supervised-300 \scriptsize (baseline) & \footnotesize 300 & \footnotesize $\mathcal{T}_{300}$ & \footnotesize 17.2\\
\hline
 \footnotesize Semi-supervised-FB & \footnotesize 910 & \scriptsize $\mathcal{T}_{300} \cup \mathcal{H}^{\textup{FB-M}}_{610}$ & \footnotesize 17.7\\
\hline
 \footnotesize Semi-supervised-STE & \footnotesize 910 & \scriptsize $\mathcal{T}_{300} \cup \mathcal{H}^{\textup{STE-M}}_{610}$ & \footnotesize 17.9\\
\hline
 \footnotesize MH-CTC (proposed) & \footnotesize 910 & \scriptsize $\mathcal{T}_{300} \cup \mathcal{H}^{\textup{FB-M}}_{610} \cup \mathcal{H}^{\textup{STE-M}}_{610}$ & \footnotesize \textbf{16.2}\\
\hline
 \footnotesize Supervised-910 & \footnotesize 910 & \footnotesize $\mathcal{T}_{910}$ & \footnotesize 6.7\\
\hline
\end{tabular}
\end{center}
\vspace{-10pt}
\end{table}
\normalsize


In both clean and multi-condition training scenarios, the supervised adaptations which use manual transcriptions for all 910 utterances have the lowest WERs. 

\section{Conclusion}
\label{sec:conclusion}

This paper has proposed an adaptation method for end-to-end speech recognition. Multiple ASR 1-best hypotheses were used in the computation of the CTC loss function to alleviate the impact of errors in the ASR hypotheses to the computation of CTC loss when the 1-best hypotheses are used as labels instead of manual transcriptions. The 1-best hypotheses were obtained by using a main ASR system and an additional ASR system which use FBANK and STE features, respectively, to decode the unlabeled adaptation data. In clean and multi-condition training scenarios, the proposed adaptation method yielded 6.6\% and 5.8\% relative WER reductions, respectively, compared to the baseline system which was adapted with back-propagation fine-tuning using an adaptation subset having manual transcriptions. In contrast, conventional semi-supervised back-propagation fine-tuning did not yield WER reduction compared to the baseline system. To our knowledge, this is the first time the integration of multiple ASR hypotheses in the CTC loss function has been shown to be consistently effective in reducing WER, and thus, is promising for future work.

\newpage 

\bibliographystyle{IEEEbib}
\bibliography{refs}

\begin{thebibliography}{10}

\bibitem{bell2020}
P.~Bell, J.~Fainberg, O.~Klejch, J.~Li, S.~Renals, and P.~Swietojanski,
\newblock ``Adaptation algorithms for speech recognition: an overview,''
\newblock in {\em arXiv preprint arXiv: 2008.06580}, 2020.

\bibitem{chorowski2015}
J.~K. Chorowski, D.~Bahdanau, D.~Serdyuk, K.~Cho, and Y.~Bengio,
\newblock ``Attention-based models for speech recognition,''
\newblock in {\em Proc. Advances in Neural Information Processing Systems
  (NIPS)}, 2015, pp. 577--585.

\bibitem{graves2014}
A.~Graves and N.~Jaitly,
\newblock ``Towards end-to-end speech recognition with recurrent neural
  networks,''
\newblock in {\em Proc. of the 31st International Conference on Machine
  Learning}, Beijing, China, June 2014, pp. 1764--1772.

\bibitem{graves2006}
A.~Graves, S.~Fernandez, F.~Gomez, and J.~Schmidhuber,
\newblock ``Connectionist temporal classification: labelling unsegmented
  sequence data with recurrent neural networks,''
\newblock in {\em Proc. of the 23rd International Conference on Machine
  Learning}, Pittsburgh, USA, June 2006, pp. 369--376.

\bibitem{paul1992}
D.~B. Paul and J.~M. Barker,
\newblock ``The design for the {W}all {S}treet {J}ournal-based {CSR} corpus,''
\newblock in {\em HLT '91 Proceedings of the workshop on Speech and Natural
  Language}, New York, USA, February 1992, pp. 357--362.

\bibitem{vincent2017}
E.~Vincent, S.~Watanabe, A.~A. Nugraha, J.~Barker, and R.~Marxer,
\newblock ``An analysis of environment, microphone and data simulation
  mismatches in robust speech recognition,''
\newblock {\em Computer Speech and Language}, vol. 46, pp. 535--557, September
  2017.

\bibitem{parihar2002}
N.~Parihar and J.~Picone,
\newblock {\em Aurora working group: {DSR} front end {LVCSR} evaluation:
  {AU}/384/02},
\newblock Institute for Signal and Information Processing Technical Report,
  2002.

\bibitem{samarakoon2018}
L.~Samarakoon, B.~Mak, and A.~Y.~S. Lam,
\newblock ``Domain adaptation of end-to-end speech recognition in low-resource
  settings,''
\newblock in {\em Proc. IEEE Spoken Language Technology Workshop}, Athens,
  Greece, December 2018, pp. 382--388.

\bibitem{li_slt2018}
K.~Li, J.~Li, Y.~Zhao, K.~Kumar, and Y.~Gong,
\newblock ``Speaker adaptation for end-to-end {CTC} models,''
\newblock in {\em Proc. IEEE Spoken Language Technology Workshop}, Athens,
  Greece, December 2018, pp. 542--549.

\bibitem{ochiai2018}
T.~Ochiai, S.~Watanabe, S.~Katagiri, T.~Hori, and J.~Hershey,
\newblock ``Speaker adaptation for multichannel end-to-end speech
  recognition,''
\newblock in {\em Proc. IEEE ICASSP}, Calgary, Canada, April 2018, pp.
  6707--6711.

\bibitem{delcroix2018}
M.~Delcroix, S.~Watanabe, A.~Ogawa, S.~Karita, and T.~Nakatani,
\newblock ``Auxilary feature based adaptation of end-to-end {ASR} systems,''
\newblock in {\em Proc. INTERSPEECH}, Hyderabad, India, September 2018, pp.
  2444--2448.

\bibitem{meng2019}
Z.~Meng, Y.~Gaur, J.~Li, and Y.~Gong,
\newblock ``Speaker adaptation for attention-based end-to-end speech
  recognition,''
\newblock in {\em Proc. INTERSPEECH}, Graz, Austria, September 2019, pp.
  241--245.

\bibitem{tsunoo2019}
E.~Tsunoo, Y.~Kashiwagi, S.~Asakawa, and T.~Kumakura,
\newblock ``End-to-end adaptation with backpropagation through {WFST} for
  on-device speech recognition system,''
\newblock in {\em Proc. INTERSPEECH}, Graz, Austria, September 2019, pp.
  764--768.

\bibitem{sari_icassp2020}
L.~Sari, N.~Moritz, T.~Hori, and J.~Le~Roux,
\newblock ``Unsupervised speaker adaptation using attention-based speaker
  memory for end-to-end {ASR},''
\newblock in {\em Proc. IEEE ICASSP}, Barcelona, Spain, May 2020, pp.
  7384--7388.

\bibitem{do_eusipco2020}
C.-T. Do, S.~Zhang, and T.~Hain,
\newblock ``Selective adaptation of end-to-end speech recognition using hybrid
  {CTC}/attention architecture for noise robustness,''
\newblock in {\em Proc. of the 28th European Signal Processing Conference
  (EUSIPCO)}, Amsterdam, The Netherlands, August 2020, pp. 321--325.

\bibitem{ding2020}
F.~Ding, W.~Guo, B.~Gu, Z.~Ling, and J.~Du,
\newblock ``Adaptive speaker normalization for {CTC}-based speech
  recognition,''
\newblock in {\em Proc. INTERSPEECH}, Shanghai, China, October 2020, pp.
  1266--1270.

\bibitem{giuliani2007}
D.~Giuliani and F.~Brugnara,
\newblock ``Experiments on cross-system acoustic model adaptation,''
\newblock in {\em Proc. IEEE Automatic Speech Recognition and Understanding
  Workshop (ASRU)}, Kyoto, Japan, Dec. 2007, pp. 117--122.

\bibitem{stueker2006}
S.~Stueker, C.~Fuegen, S.~Burger, and M.~Woelfel,
\newblock ``Cross-system adaptation and combination for continuous speech
  recognition: the influence of phoneme set and acoustic front-end,''
\newblock in {\em Proc. INTERSPEECH}, Pittsburgh, USA, September 2006, pp.
  521--524.

\bibitem{lecun1995}
Y.~LeCun and Y.~Bengio,
\newblock ``Convolutional networks for images, speech, and time series,''
\newblock in {\em The Handbook of Brain Theory and Neural Networks}. MIT Press,
  1995.

\bibitem{hochreiter1997}
S.~Hochreiter and J.~Schmidhuber,
\newblock ``Long short-term memory,''
\newblock {\em Neural Computation}, vol. 9, pp. 1735--1780, November 1997.

\bibitem{rumelhart1986}
D.~E. Rumelhart, G.~E. Hinton, and R.~J. Williams,
\newblock ``Learning representations by back-propagating errors,''
\newblock {\em Nature}, vol. 323, no. 9, pp. 533--536, 1986.

\bibitem{ruder2016}
S.~Ruder,
\newblock ``An overview of gradient descent optimisation algorithms,''
\newblock in {\em arXiv preprint arXiv: 1609.04747}, 2016.

\bibitem{mohamed2012}
A.-R. Mohamed, G.~Hinton, and G.~Penn,
\newblock ``Understanding how deep belief networks perform acoustic
  modelling,''
\newblock in {\em Proc. IEEE ICASSP}, Kyoto, Japan, March 2012, pp. 4273--4276.

\bibitem{watanabe2018}
S.~Watanabe, T.~Hori, S.~Karita, T.~Hayashi, J.~Nishitoba, Y.~Unno, N.~E.~Y.
  Soplin, J.~Heymann, M.~Wiesner, N.~Chen, A.~Renduchintala, and T.~Ochiai,
\newblock ``{ESP}net: end-to-end speech processing toolkit,''
\newblock in {\em Proc. INTERSPEECH}, Hyderabad, India, September 2018, pp.
  2207--2211.

\bibitem{povey2011}
D.~Povey, A.~Ghoshal, G.~Boulianne, L.~Burget, O.~Glembek, N.~Goel,
  M.~Hannemann, P.~Motlicek, Y.~Qian, P.~Schwarz, J.~Silovsky, G.~Stemmer, and
  K.~Vesely,
\newblock ``The {K}aldi speech recognition toolkit,''
\newblock in {\em Proc. IEEE Automatic Speech Recognition and Understanding
  Workshop (ASRU)}, Hawaii, USA, December 2011.

\bibitem{do2017_interspeech}
C.-T. Do and Y.~Stylianou,
\newblock ``Improved automatic speech recognition using subband temporal
  envelope features and time-delay neural network denoising autoencoder,''
\newblock in {\em Proc. INTERSPEECH}, Stockholm, Sweden, August 2017, pp.
  3832--3836.

\bibitem{doddipatla2018}
R.~Doddipatla, T.~Kagoshima, C.-T. Do, P.N. Petkov, C.~Zorila, E.~Kim,
  D.~Hayakawa, H.~Fujimura, and Y.~Stylianou,
\newblock ``The {T}oshiba entry to the {CH}i{ME} 2018 challenge,''
\newblock in {\em Proc. CHiME 2018 Workshop on Speech Processing in Everyday
  Environments}, Hyderabad, India, September 2018, pp. 41--45.

\bibitem{do2019}
C.-T. Do,
\newblock ``Subband temporal envelope features and data augmentation for
  end-to-end recognition of distant conversational speech,''
\newblock in {\em Proc. IEEE ICASSP}, Brighton, UK, May 2019, pp. 6251--6255.

\bibitem{simonyan2015}
K.~Simonyan and A.~Zisserman,
\newblock ``Very deep convolutional networks for large-scale image
  recognition,''
\newblock in {\em Proc. International Conference on Learning Representations},
  2015.

\end{thebibliography}

\end{document}